\documentclass[10pt,twocolumn,letterpaper]{article}

 \usepackage{cvpr}              

\usepackage{multicol}
\usepackage{dblfloatfix}
\usepackage{graphicx} 
\usepackage{makecell} 
\usepackage{float}
\usepackage{multirow}
\usepackage[accsupp]{axessibility}

\newcommand\blfootnote[1]{%
  \begingroup
  \renewcommand\thefootnote{}\footnote{#1}%
  \addtocounter{footnote}{-1}%
  \endgroup
}

\usepackage[dvipsnames]{xcolor}

\definecolor{cvprblue}{rgb}{0.21,0.49,0.74}
\usepackage[pagebackref,breaklinks,colorlinks,citecolor=cvprblue]{hyperref}

\def\paperID{####} 
\def\confName{CVPR}
\def\confYear{2024}

\title{Zero-Shot Dual-Path Integration Framework for Open-Vocabulary 3D Instance Segmentation}

\author{Tri Ton\(^{1*}\), Ji Woo Hong\(^{1*}\), SooHwan Eom\(^{1}\), Jun Yeop Shim\(^{1}\), Junyeong Kim\(^{2}\), Chang D. Yoo\(^{1}\) \\
\(^{1}\)Korea Advanced Institute of Science and Technology (KAIST)\, \(^{2}\)Chung-Ang University \\
{\tt\small \{tritth, jiwoohong93, sean1105, shimjay17, cd\_yoo\}@kaist.ac.kr, 
junyeongkim@cau.ac.kr}
}

\begin{document}

\twocolumn[{
\maketitle
\begin{center}
   \includegraphics[width=1.0\textwidth]{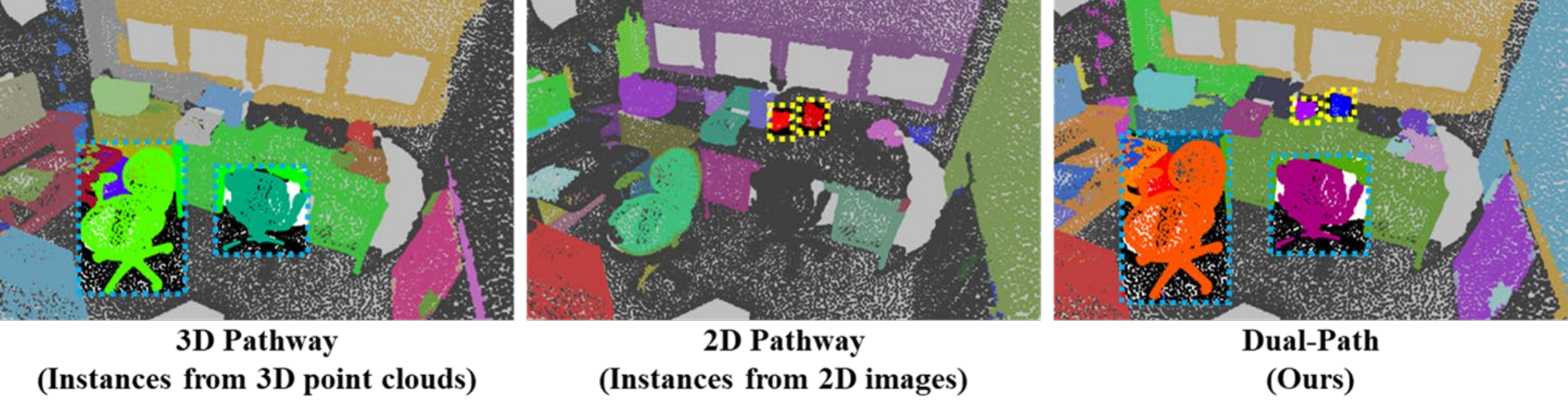}
   \captionof{figure}{Instance segmentation results from different modality of 2D and 3D. Our Zero-Shot Dual-Path Integration Framework complementarily integrates outputs from two modalities.}
\label{fig:task}
\end{center}
}]

\blfootnote{*equal contribution.

This work was supported by Institute for Information \& communications Technology Planning \& Evaluation (IITP) grant funded by the Korea government(MSIT) (No. 2021-0-01381, Development of Causal AI through Video Understanding and Reinforcement Learning, and Its Applications to Real Environments) and (No.2022-0-00184, Development and Study of AI Technologies to Inexpensively Conform to Evolving Policy on Ethics).}

\begin{abstract}
\label{sec:abstract}
Open-vocabulary 3D instance segmentation transcends traditional closed-vocabulary methods by enabling the identification of both previously seen and unseen objects in real-world scenarios. It leverages a dual-modality approach, utilizing both 3D point clouds and 2D multi-view images to generate class-agnostic object mask proposals. Previous efforts predominantly focused on enhancing 3D mask proposal models; consequently, the information that could come from 2D association to 3D was not fully exploited. This bias towards 3D data, while effective for familiar indoor objects, limits the system's adaptability to new and varied object types, where 2D models offer greater utility. Addressing this gap, we introduce Zero-Shot Dual-Path Integration Framework that equally values the contributions of both 3D and 2D modalities. Our framework comprises three components: 3D pathway, 2D pathway, and Dual-Path Integration. 3D pathway generates spatially accurate class-agnostic mask proposals of common indoor objects from 3D point cloud data using a pre-trained 3D model, while 2D pathway utilizes pre-trained open-vocabulary instance segmentation model to identify a diverse array of object proposals from multi-view RGB-D images. In Dual-Path Integration, our Conditional Integration process, which operates in two stages, filters and merges the proposals from both pathways adaptively. This process harmonizes output proposals to enhance segmentation capabilities. Our framework, utilizing pre-trained models in a zero-shot manner, is model-agnostic and demonstrates superior performance on both seen and unseen data, as evidenced by comprehensive evaluations on the ScanNet200 and qualitative results on ARKitScenes datasets.
\end{abstract}
\section{Introduction}
\label{sec:intro}

The advent of 3D instance segmentation task, which predicts 3D object instances and their corresponding categories from 3D point cloud data, has marked a significant milestone for its applications in autonomous driving, robotics, augmented reality, and more. 
Traditional 3D instance segmentation methods \cite{hou2019sis, yang2019learning, Yi2018GSPNGS, Yi2018GSPNGS, jiang2020pointgroup, Chen_HAIS_2021_ICCV, Engelmann20CVPR, liang2021instance, He2021dyco3d, wu20223d, vu2022softgroup, ngo2023isbnet, zhao2023divide, Schult23ICRA, 2211.15766} have largely operated within a closed-vocabulary paradigm, where the classes of objects to be segmented are predetermined and known during the training phase. 
However, real-world scenarios frequently present objects that fall outside of these predefined classes, making closed-vocabulary segmentation inadequate.
Open-vocabulary 3D instance segmentation emerged as a necessary evolution to address these real-world complexities. 
The challenge lies in the system's ability to generalize beyond the training dataset and to adapt to the unpredictability inherent in real-world environments.

\begin{figure}[t] 
\centering
\includegraphics[width=\columnwidth]{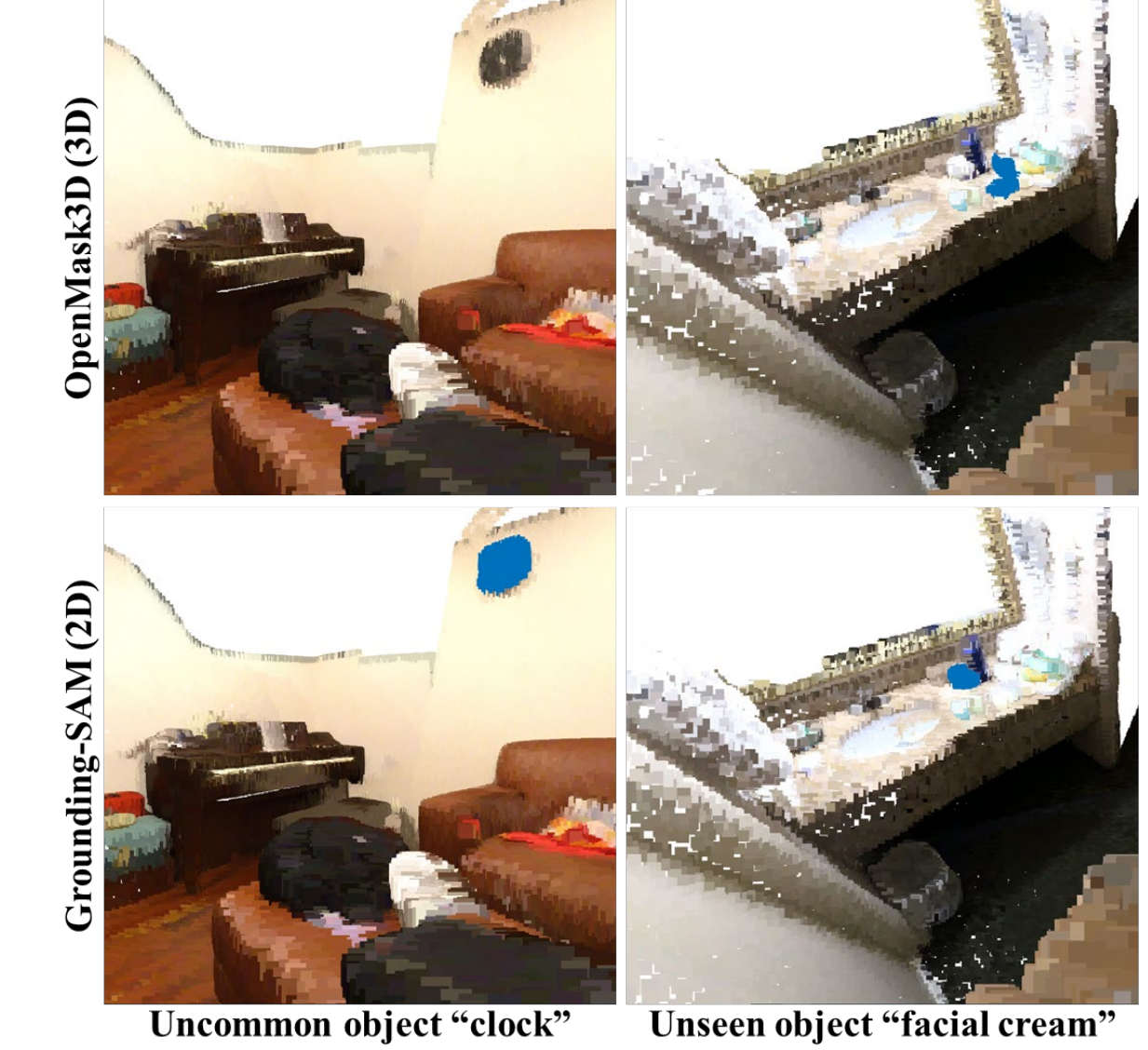} 
\caption{Capability of pre-trained open-vocabulary 2D instance segmentation in detecting uncommon and unseen object classes that remain undetected by pre-trained 3D instance segmentation models.}
\label{fig:prob}
\vskip -0.15in
\end{figure}

Current methodologies \cite{takmaz2023openmask3d, huang2023openins3d} independently process 3D and 2D data of the same scene for discrete sub-tasks without harnessing the inherent synergistic potential between two modalities and heavily depend on the initial 3D instance masks for subsequent classification.
From these works, it has been observed that while a class-agnostic 3D instance segmentation model using point cloud data excels at segmenting common objects such as `TV', it often struggles with uncommon classes like `paper towel roll' and unseen classes like `facial cream'. 
Conversely, open-vocabulary 2D instance segmentation excels at detecting unfamiliar objects within 3D data, benefiting from the robust generalization afforded by the vision-language understanding capabilities of pre-trained image classifiers, as illustrated in Fig. \ref{fig:prob}.
Thus, we argue that leveraging both modalities of the same scene offers distinct advantages.

Bridging the gap between two modalities, we propose a Zero-shot Dual-Path Integration Framework designed to synergistically merge the instance proposals derived from the 3D point cloud and the 2D multi-view images, which excel in spatial precision on common objects and diverse recognition capabilities respectively. Our framework comprises three components: 3D pathway, 2D pathway, and Dual-Path Integration. 

In the 3D pathway, 3D instance masks are generated from a pre-trained class-agnostic mask proposal generator using a 3D point cloud of spatial information to generate as many proposals as possible. 
In the 2D pathway, 2D instance masks are generated from an pre-trained open vocabulary 2D instance segmentation using RGB-D multi-view image sequence of 2D visual details, which are then projected into the 3D scan and refined through Instance Fusion Module.

Employing a straightforward integration approach of using all proposals from both the 3D and 2D pathways, as we call Simple Integration, will enhance the performance of instance segmentation to a certain extent as the diversity of proposals enhances the recall rate.
Yet, the quality of the proposals is also critical, as integrating numerous but low-quality proposals can lead to increased false positive, which may adversely affect the overall precision. Consequently, while the diversity of proposals can boost the recall, it is essential to maintain a balance between quantity and quality. 
To this end, in the final Dual-Path Integration phase, our meticulously designed Conditional Integration evaluates proposals from both the 3D and 2D modalities with impartial consideration. This module unfolds in two pivotal stages: (1) Dual-modality Proposal Matching and (2) Adaptive Integration. 
During the Dual-modality Proposal Matching stage, we compute Intersection-of-Union (IoU) across the proposals from 3D and 2D pathways. This aims to identify pairs of proposals with overlapping regions, suggesting they may represent parts of identical objects. 
The unique proposals with no overlap with any other proposals of other modality are filtered to be added to the final proposal outputs.
Subsequently, Adaptive Integration establishes a balance between segmentation precision and the identification of diverse instances based on the IoU assessment.
The samples of the 3D pathway, 2D pathway, and Dual-path instances can be seen in Fig. \ref{fig:task}.

Our contributions are as follows: 
\begin{itemize}
\item \textbf{Zero-shot 3D instance segmentation via comprehensive exploitation of pre-trained models of both 3D and 2D modalities}: We introduce zero-shot framework that judiciously leverages the strengths of pre-trained models of both 3D point cloud and 2D image data for 3D instance segmentation. This approach embraces a model-agnostic strategy that avoids the traditional dependency on a pre-trained model of single modality.
\item \textbf{Dual-Path Integration of Conditional Integration}: We propose a Dual-Path Integration with Conditional Integration process that effectively combines instance mask proposals from both 3D and 2D pathways to reconcile and enhance mask proposals.
\item \textbf{Enhanced Overall Performance}: Our framework's efficacy is validated through evaluations on the ScanNet200 and qualitative results in ARKitScenes. The results demonstrate an uplift in the overall performance for open-vocabulary 3D instance segmentation, underscoring the potency of our proposed approach.
\end{itemize}

\begin{figure*}[t]
\centering
\includegraphics[width=\textwidth]{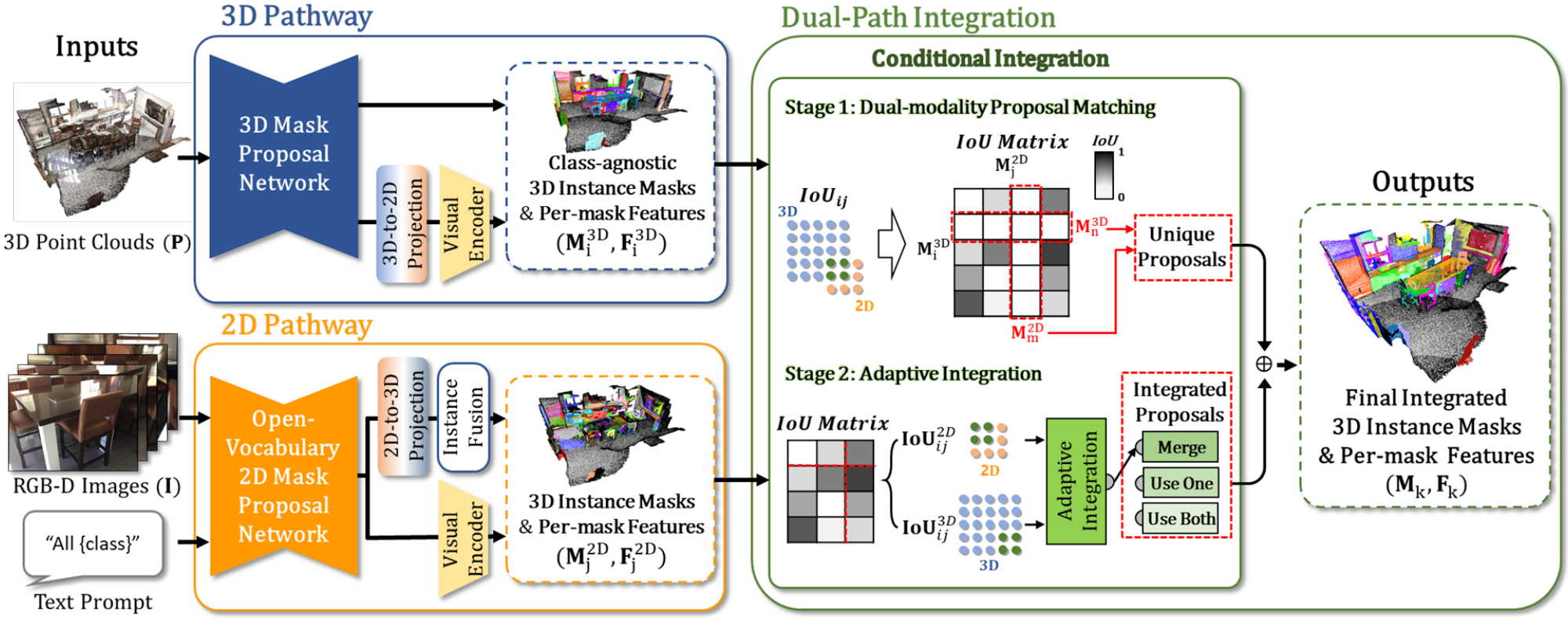} 
\caption{Overview of our Zero-Shot Dual-Path Integration Framework. The 3D pathway takes 3D point cloud $\mathbf{P}$ as input to generated class-agnostic 3D instance masks $\mathbf{M}_\text{i}^\text{3D}$ with pre-trained 3D Mask Proposal Network and the per-mask visual features $\mathbf{F}_\text{i}^\text{3D}$ are extracted with CLIP visual encoder \cite{radford2021learning}. The 2D pathway also generates its own 3D instance masks $\mathbf{M}_\text{j}^\text{2D}$ using RGB-D Image $\mathbf{I}$ input with Open-vocabulary 2D Mask Proposal Network and 2D-to-3D Projection module, along with the per-mask visual features $\mathbf{F}_\text{i}^\text{2D}$ of each mask. The outputs of two pathways are integrated through the Conditional Integration which utilizes Intersection-of-Union (IoU) for Dual-modality Proposal Matching and Adaptive Integration, having final 3D instance results $\mathbf{M}_\text{k}$ and their visual features $\mathbf{F}_\text{k}$ as outputs.}
\label{fig:overview}
\vskip -0.15in
\end{figure*}
\section{Related Work}
\label{sec:related}

\paragraph{Closed Vocabulary 3D Instance Segmentation.}

3D instance segmentation techniques are categorized into proposal-based, clustering-based, and transformer-based approaches. Proposal-based methods \cite{hou2019sis, yang2019learning, Yi2018GSPNGS} identify 3D bounding boxes to distinguish instances, yet face challenges with varying point cloud distributions. Clustering-based approaches \cite{Yi2018GSPNGS, jiang2020pointgroup, Chen_HAIS_2021_ICCV, Engelmann20CVPR, liang2021instance, He2021dyco3d, wu20223d, vu2022softgroup, ngo2023isbnet, zhao2023divide} predict semantic categories and use geometric offsets for instance grouping, though they require extensive manual tuning and may struggle with novel test objects. Transformer-based methods \cite{Schult23ICRA, 2211.15766}, leveraging the Mask2Former framework \cite{cheng2021mask2former}, achieve state-of-the-art results by representing instances as queries within a transformer decoder, effectively utilizing global features for mask prediction, demonstrating robust performance, particularly on the ScanNet benchmark \cite{dai2017scannet}.

\paragraph{Open Vocabulary 2D Instance Segmentation.}

Open-vocabulary 2D segmentation, empowered by large-scale vision-language models such as CLIP \cite{radford2021learning}, LiT \cite{zhai2022lit}, and ALIGN \cite{jia2021scaling}, has emerged as a significant advancement in instance segmentation. This development has fostered novel open-vocabulary and zero-shot segmentation methodologies. Pixel-level embedding techniques \cite{ghiasi2022scaling, liang2023open, xu2022groupvit, huynh2022open, wu2023betrayed, vs2023mask} have shown promising results, although their success is contingent on mask precision and requires specific training. To overcome these challenges, integrating robust zero-shot detection and segmentation models such as ViLD \cite{gu2021open}, OWL-ViT \cite{minderer2205simple}, Detic \cite{zhou2022detecting}, and Grounding-DINO \cite{liu2023grounding} with the Segment Anything Model (SAM) \cite{kirillov2023segment} facilitates open-vocabulary 2D instance segmentation. This approach, especially when combining Grounding-DINO with SAM (Grounded-SAM), leverages CLIP features for classification, enabling precise segmentation and classification without the necessity for additional fine-tuning or training. In our work, we employ Grounded-SAM for open-vocabulary 2D instance segmentation.

\paragraph{Open Vocabulary 3D Scene Understanding.}

Recent advancements in open-vocabulary 3D scene understanding \cite{ding2022language, ha2022semabs, huang23vlmaps, conceptfusion, lerf2023, peng2023openscene, takmaz2023openmask3d, lu2023ovir, huang2023openins3d, conceptgraphs} have focused on leveraging 2D vision-language model features for 3D reconstruction due to the absence of large-scale 3D datasets. Techniques like OpenScene \cite{peng2023openscene} and ConceptFusion \cite{conceptfusion} utilize pixel-wise CLIP features \cite{radford2021learning} for text-aligned 3D feature extraction. LERF \cite{lerf2023} employs similar CLIP-based \cite{radford2021learning} semantic fields within NeRF \cite{mildenhall2020nerf} frameworks, providing query-specific scene heatmaps but limited object instance understanding. Instance-based methodologies \cite{yang2023sam3d, takmaz2023openmask3d, lu2023ovir, huang2023openins3d, conceptgraphs} like SAM3D \cite{yang2023sam3d} and OpenMask3D \cite{takmaz2023openmask3d}, project 2D masks onto 3D point clouds for enhanced instance recognition, though challenges remain in instance merging and quality of 3D mask proposals. OVIR-3D \cite{lu2023ovir} attempts to improve instance merging but struggles with large instance backgrounds. Our method overcomes these obstacles by integrating high-performing pre-trained models from both 2D and 3D domains, reducing reliance on single-modality insights.

\section{Method}
\label{sec:method}

The Zero-Shot Dual-Path Integration Framework is designed to predict class-agnostic 3D instance masks and their corresponding CLIP-based features for open-vocabulary instance classification, as illustrated in Fig. \ref{fig:overview}. This framework operates on posed RGB-D images and reconstructed 3D point clouds, leveraging queries to isolate class-specific instance masks through a dual-pathway approach: a 3D pathway for mask proposal generation with point clouds and a 2D pathway for RGB-D based mask proposal generation.


In our approach, the 3D pathway employs a mask proposal network to generate 3D instance masks, integrating 2D bounding boxes and CLIP-based features to bridge open-vocabulary concepts with these masks, leveraging architectures adept at detecting sizeable objects \cite{Schult23ICRA}. Simultaneously, the 2D pathway applies an open-vocabulary 2D instance segmentation network to RGB-D images, creating 2D mask proposals. These are then projected into the 3D point cloud, refined into finalized instance masks by an Instance Fusion Module.


Integration of these pathways is achieved through our Conditional Integration in the Dual-Path Integration phase, which encompasses Dual-modality Proposal Matching and Adaptive Integration. This process begins with the matching of overlapping proposals from both pathways using Intersection-of-Union (IoU) metrics, followed by the conditional merging of these proposals based on the assessment of their IoU. This integration technique enhances the framework's accuracy in object recognition and segmentation by leveraging the complementary strengths of both 2D and 3D data.

\subsection{3D Pathway: 3D Mask Proposals from Point Cloud}
\label{sec:method_3d}

Given a 3D point cloud, denoted as \( \mathbf{P} \in \mathbb{R}^{N \times 3} \) where \( N \) represents the total number of points, each point is represented by a 3D position. 
The 3D pathway's objective is to segment the given point cloud into class-agnostic 3D instance mask proposals, represented through a collection of binary masks, where each mask is denoted as \( \mathbf{M}_\text{i}^\text{3D} = ( {M}_\text{i,1}^\text{3D}, ..., {M}_\text{i,N}^\text{3D} ) \) where \( {M}_\text{i,n}^\text{3D} \in \{0, 1\} \), meaning \(n\)-th point belongs to \(i\)-th object instance. 
For generating the 3D instance masks, Mask3D \cite{Schult23ICRA} is used as our 3D instance segmentation network, which utilizes U-Net style sparse convolutional backbone \cite{graham20183d} as a feature extractor. A fixed number of object queries go through the transformer decoder layers to attend to global features iteratively, directly outputting instance predictions. This generates a binary mask for each instance with predicted class labels and their confidence scores. However, since we want a class-agnostic mask proposal network, Mask3D \cite{Schult23ICRA} is modified to omit the predicted class labels and confidence scores to only concentrate on generating binary instance mask proposals. This way, we obtain open-vocabulary representations instead of semantic class predictions confined to closed-vocabulary. 

Then, we derive per-mask text-aligned visual features using pre-trained CLIP \cite{hegde2023clip} for querying open-vocabulary concepts associated with predicted instance masks. Inspired by \cite{takmaz2023openmask3d}, we first select the top $k$ RGB-D images with the highest visibility of each instance mask. In prior work \cite{takmaz2023openmask3d}, points projected onto the images are utilized as the prompts to guide the SAM \cite{kirillov2023segment} in generating the 2D bounding boxes, which are utilized for cropping the images for CLIP feature extraction. However, this projection can cause errors due to approximation in the occlusion test, and the random selection of the $k$ projected points might erroneously include points lying outside the actual instances, leading to the generation of poor-quality bounding boxes.
Consequently, the extracted CLIP features from those poor-quality bounding boxes suffer from inadequacies. To bypass this issue, we directly use the projected 3D bounding box into 2D images to get CLIP-based features, eliminating potential errors in random point selection. 
We employ multi-level crops of specific regions for feature enrichment to encapsulate extensive contextual details from the surrounding environment. Leveraging the CLIP visual encoder \cite{radford2021learning}, the CLIP feature vectors \( \mathbf{F}_\text{i}^\text{3D} \in \mathbb{R}^{d} \) of the cropped object images are extracted and average-pooled to generate final mask-feature representations for each object.

\subsection{2D Pathway: 3D Mask Proposals from Multi-view RGB-D Images}
The objective of 2D pathway is to generate 3D instance masks, where each mask is denoted as \( \mathbf{M}_\text{j}^\text{2D} = ( {M}_\text{j,1}^\text{2D}, ..., {M}_\text{j,N}^\text{2D} ) \), where \( {M}_\text{j,n}^\text{2D} \in \{0, 1\} \), meaning \(n\)-th point belongs to \(j\)-th object instance, from RBG-D image \( \mathbf{I}_\text{t} \) where \( t \) is the image frame at time \( t \) with known camera intrinsic matrix \( C \) and world-to-camera extrinsic matrix (pose) \( E_{t} \).
For this process, we first utilize Grounded-SAM, a fusion of Grounding DINO \cite{liu2023grounding} and SAM \cite{kirillov2023segment}, as pre-trained 2D open-vocabulary instance segmentation network to obtain 2D mask proposals \( \mathbf{m}_\text{t,j}^\text{2D} \). 
Grounding DINO takes the text prompt as an input to produce the 2D bounding boxes, which SAM subsequently uses to obtain 2D mask proposals. These 2D mask proposals \( \mathbf{m}_{t,j}^{2D} \) undergo subsequent projection into the 3D point cloud with known camera intrinsic, pose, and depth. 
Additionally, we extract CLIP-based features \( \mathbf{F}^\text{2D}_\text{j} \in \mathbb{R}^{d} \) from the corresponding cropped image for each proposal. 

Given that mask proposals sourced from 2D images may be fragmented due to occlusion, these proposals from the 2D pathway are passed to an Instance Fusion process for complete proposals. Drawing upon methodologies from \cite{lu2023ovir}, the Instance Fusion Module accumulates 3D projected instances within a memory bank. It then periodically executes a filtering and merging process, utilizing the 3D Intersection-of-Union metric in conjunction with feature similarity analysis.

\subsection{Dual-Path Integration}
Our Dual-path Integration framework incorporates Conditional Integration that operates through a meticulously structured two-stage process. 
First, Dual-modality Proposal Matching stage utilizes the Intersection-of-Union (IoU) metrics to effectively identify unique proposals from each respective modality. 
Given 3D instances from 3D point cloud \( \mathbf{M}_\text{i}^\text{3D} \) and 2D multi-view images \( \mathbf{M}_\text{j}^\text{2D} \) where \(i\) and \(j\) are the number of generated instances for each pathway, the module systematically calculates the Intersection-of-Union (IoU), denoted as \( \mathbf{IoU}_{ij} \) as Equation \ref{eq:1}: 
\begin{equation}
\mathbf{IoU}_{ij} = \frac{|\mathbf{M}_\text{i}^\text{3D} \cap \mathbf{M}_\text{j}^\text{2D}|}{|\mathbf{M}_\text{i}^\text{3D} \cup \mathbf{M}_\text{j}^\text{2D}|.}
\label{eq:1}
\end{equation}
This computation is conducted for each possible pair of instances across the modalities.
As a result, IoU matrix providing a comprehensive representation of the spatial relationships between all instances across the 3D and 2D modalities is created.

To identify unique proposals from each modality for inclusion in the final instance proposals \( \mathbf{M}_\text{k} \), we systematically evaluate instances from \( \mathbf{M}_\text{j}^\text{2D} \) and \( \mathbf{M}_\text{i}^\text{3D} \) against the IoU matrix. This process aims to detect instances without overlap across the entirety of instances from the alternate modality. This evaluation can be articulated as follows:
\begin{align*}
\forall j, &\, \text{if} \, (\forall i, \, \mathbf{IoU}_{ij} = 0), \, \text{then add} \, \mathbf{M}_\text{j}^\text{2D} \, \text{to} \, \mathbf{M}_\text{k} \\
\forall i, &\, \text{if} \, (\forall j, \, \mathbf{IoU}_{ij} = 0), \, \text{then add} \, \mathbf{M}_\text{i}^\text{3D} \, \text{to} \, \mathbf{M}_\text{k}.
\end{align*}
This procedure ensures that any instance \( \mathbf{M}_\text{j}^\text{2D} \) lacking overlap with all \( \mathbf{M}_\text{i}^\text{3D} \) instances (indicated by an \( \mathbf{IoU}_{ij} \) value of 0) is directly incorporated into \( \mathbf{M}_\text{k} \). Conversely, it also guarantees the inclusion of any \( \mathbf{M}_\text{i}^\text{3D} \) instance that does not overlap with all \( \mathbf{M}_\text{j}^\text{2D} \) instances into \( \mathbf{M}_\text{k} \). 
Additionally, proposal pairs exhibiting the smallest \( \mathbf{IoU}_{ij} \) for all instances \( \mathbf{M}_\text{j}^\text{2D} \) are added to \( \mathbf{M}_\text{k} \), aiming to enrich the segmentation with a broader array of detected objects. 

Excluding the unique instances identified in the first stage above, our Conditional Integration progresses to the Adaptive Integration stage, where it performs additional computation of IoUs bifurcated into two distinct perspectives, as in Equations \ref{eq:2}:
\begin{align}
\mathbf{IoU}^{3D}_{ij} = \frac{|\mathbf{M}_\text{i}^\text{3D} \cap \mathbf{M}_\text{j}^\text{2D}|}{|\mathbf{M}_\text{i}^\text{3D}|}, \quad
\mathbf{IoU}^{2D}_{ij} = \frac{|\mathbf{M}_\text{i}^\text{3D} \cap \mathbf{M}_\text{j}^\text{2D}|}{|\mathbf{M}_\text{j}^\text{2D}|.}
\label{eq:2}
\end{align}
Equation \ref{eq:2} (left) introduces \( \mathbf{IoU}^{3D}_{ij} \), which quantifies the proportion of overlap between an instance from 3D pathway \( \mathbf{M}_\text{i}^\text{3D} \) and a instance from 3D pathway \( \mathbf{M}_\text{i}^\text{3D} \) relative to the entire instance from 3D pathway. 
Conversely, Equation \ref{eq:2} (right) defines \( \mathbf{IoU}^{2D}_{ij} \) as the ratio of their intersection to the total area of the instance from 2D pathway\( \mathbf{M}_\text{j}^\text{2D} \). 
Together, these IoU metrics offer a dual perspective on the relations between pairs of instances from the 3D pathway and the 2D pathway.
By analyzing the extent to which an instance from one pathway encompasses the spatial domain of the instance from another pathway, it offers insights into the priority between two proposals.

Subsequently, we select the proposal pair with highest \( \mathbf{IoU}_{ij} \) for each proposals of \( \mathbf{M}_\text{j}^\text{2D} \) and assess them into four scenarios for Adaptive Integration: (1) high IoU for both \( \mathbf{IoU}^{2D}_{ij} \) and \( \mathbf{IoU}^{3D}_{ij} \), (2) low IoU for both, (3) high IoU for the \( \mathbf{IoU}^{2D}_{ij} \) but low for \( \mathbf{IoU}^{3D}_{ij} \) (e.g. proposal from the 2D pathway is subgroup of that from the 3D pathway), and (4) the vice-versa. 
By selecting the proposal pair exhibiting the highest \( \mathbf{IoU}_{ij} \), we select the proposal from the alternate pathway that demonstrates the most significant relationship, encapsulating both concordance and discordance, for further evaluation.

The four distinct scenarios of Adaptive Integration are elaborated below:
\begin{enumerate}
    \item \textbf{Significant overlap}: For proposal pairs exhibiting extensive overlap, it is inferred that they likely depict the same object. Thus, these proposals are merged into a singular, comprehensive proposal for inclusion in \( \mathbf{M}_\text{k} \), ensuring a unified representation.
    
    \item \textbf{Slight overlap}: When a pair demonstrates only slight overlap, yet selected beforehand for having maximum overlap among all considered pairs, it is surmised that the proposals likely denote two distinct objects in close proximity. Accordingly, both proposals are maintained separately in \( \mathbf{M}_\text{k} \), preserving the individuality of each detected object.
    
    \item \textbf{Proposal from 2D is subgroup of proposal from 3D}: In instances where a proposal from 2D pathway is almost entirely encompassed by a proposal from 3D pathway, it is treated as a unique finding exclusive to the 2D pathway and thus given precedence for inclusion into \( \mathbf{M}_\text{k} \). This decision is based on the assumption that the proposal from 2D modality may highlight a detail or aspect not captured from the 3D modality. Although the larger overlapping proposal is neglected in this instance, its potential value is recognized. We anticipate that it will align with nearby 2D pathway proposals, thereby being considered under different scenarios.
    
    \item \textbf{Proposal from 3D is subgroup of proposal from 2D}: Analogous to scenario (3) with reversed roles, the proposal from 3D pathway is prioritized for addition to \( \mathbf{M}_\text{k} \). This reflects the broader spatial coverage and potentially significant detection afforded by the 3D pathway.
\end{enumerate}

Through these scenario-specific strategies, our integration process adeptly balances the quantity and quality of proposals offered by both pathways, enhancing the overall accuracy and completeness of instance segmentation.
To differentiate between 'high' and 'low' IoU values, thresholds for each pathway are designated as  $\theta_{2D}$ and $\theta_{3D}$, respectively, with their optimal values determined through empirical experimentation. 
Leveraging visual features derived from 3D point clouds and 2D multi-view images, \( \mathbf{F}_\text{i}^\text{3D} \) and \( \mathbf{F}_\text{j}^\text{2D} \), we employ an averaging process when merging two masks.
After the Adaptive Integration, we obtain the final instance proposals \( \mathbf{M}_\text{k} \) and the corresponding CLIP-based features \( \mathbf{F}_\text{k} \) ready to perform a semantic label assignment.

During the inference phase, a textual query denoted as $q$ correlates with a repository of representative features linked to individual 3D instances. The text feature $\mathbf{F}_\text{q}$ will be extracted from the CLIP encoder to compare with instance features $\mathbf{F}_\text{k}$ in the scene. Subsequently, these instances denoted as $\mathbf{M}_\text{k}$, undergo a ranking process based on their resemblance to the query. The top-ranked instances, thus determined, are retrieved and subsequently returned.

\begin{table*}[h]
\centering
\begin{tabular}{lcccccc}
\Xhline{2\arrayrulewidth}
Model & AP $\uparrow$ & $\text{AP}_{50}$ $\uparrow$ & $\text{AP}_{25}$ $\uparrow$ & head (AP) $\uparrow$ & common (AP) $\uparrow$ & tail (AP) $\uparrow$ \\ \Xhline{2\arrayrulewidth}
SAM3D \cite{yang2023sam3d}  & 6.1 & 14.2   & 21.3   & 7.0       & 6.2         & 4.6         \\ 
OpenScene \cite{peng2023openscene}  & 11.7 & 15.2   & 17.8   & 13.4       & 11.6         & 9.9       \\
OVIR-3D \cite{lu2023ovir}  & 13.0 & 24.9   & \textbf{32.3}   & 14.4       & 12.7         & 11.7         \\
OpenMask3D \cite{takmaz2023openmask3d}  & 15.4 & 19.9   & 23.1   & 17.1       & 14.1         & \textbf{14.9}       \\
Dual-Path (Ours)   & \textbf{19.9} & \textbf{25.0}   & 27.1   & \textbf{27.9}       & \textbf{18.9}         & 11.5       \\ \Xhline{2\arrayrulewidth}
\end{tabular}
\caption{Evaluation of Open-vocabulary 3D instance segmentation on the 312 scenes of ScanNet200 \cite{dai2017scannet} validation set. Average Precision (AP) measured over a range of overlap thresholds, 50\% overlaps, and 25\% overlaps. AP of ``head'', ``common'', and ``tail'' subsets \cite{rozenberszki2022language} of ScanNet200  are also reported.}
\label{table:quan}
\end{table*}
\section{Experiments}
\label{sec:experiments}


In this section, we present both quantitative and qualitative outcomes of our Zero-shot Dual-Path Integration Framework. We conduct a quantitative assessment, comparing our open-vocabulary 3D instance segmentation method against existing approaches within a closed-vocabulary setting. Ablation studies further dissect the impact of baseline architectures for each pathway, alongside the examination of metrics and thresholds pivotal to our Dual-path Integration process. Additionally, we showcase qualitative results from the ScanNet200 and ARKitScenes datasets to underscore our method's effectiveness in open-vocabulary 3D instance segmentation, demonstrating its proficiency in accurately identifying a wide spectrum of objects.

\subsection{Dataset and Metric}
\paragraph{Dataset.}


Our evaluation framework employs the ScanNet200 benchmark dataset \cite{dai2017scannet}, utilizing its validation set of 312 unique scenes for 3D instance segmentation performance assessment across a closed vocabulary of 200 categories. Further, we adopt the categorization scheme by Rozenberszki et al. \cite{rozenberszki2022language}, dividing the ScanNet200 object classes into ``head'', ``common'', and ``tail'' subsets, with 66, 68, and 66 categories respectively, to analyze model performance across varying object occurrence frequencies.
Additionally, we leverage the ARKitScenes dataset \cite{baruch2021arkitscenes}, which consists of over 5K scans from approximately 1.6K diverse indoor settings, offering 3D mesh reconstructions, RGB and depth images, and ARKitSLAM-estimated camera poses. This dataset aids in simulating realistic indoor scanning trajectories. Performance analysis is further enhanced by employing queries from OpenSUN3D \cite{engelmann2024opensun3d} within the Challenge development set, demonstrating the effectiveness of our approach in advanced indoor scene understanding tasks.

\paragraph{Metric.}
In our evaluation, we adopt the widely recognized metric for 3D instance segmentation: average precision (AP). 
The AP scores are calculated at mask overlap thresholds of 50\%, 25\%, and average over the overlap range of [0.5 : 0.95 : 0.05], in line with the ScanNet \cite{dai2017scannet} evaluation protocol. 
Furthermore, we analyze the AP scores across the ``head'', ``common'', and ``tail'' subsets of ScanNet200. 
This allows us to gain deeper insights into the performance of our method across different frequency categories.

\begin{figure}[ht!] 
\centering
\includegraphics[width=\columnwidth]{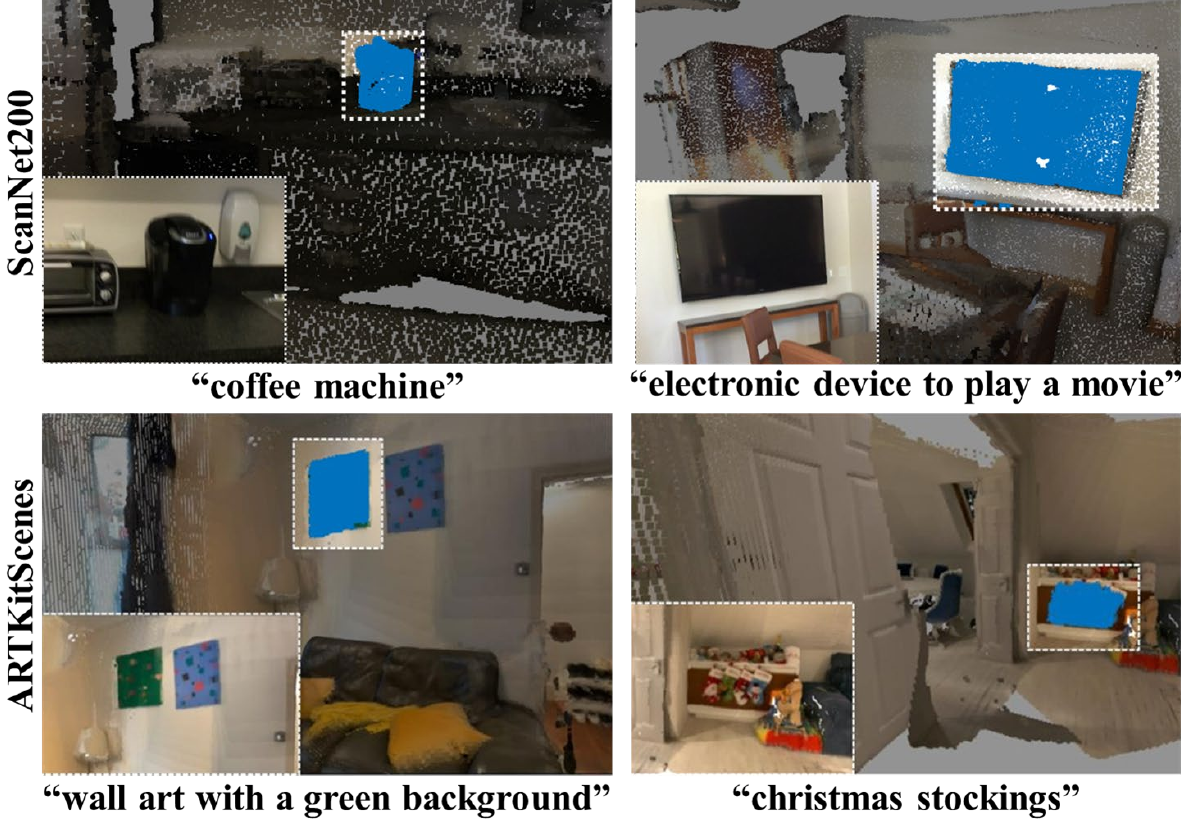} 
\caption{Qualitative results showcasing the proficiency of our framework in performing open-vocabulary 3D instance segmentation. The displayed results include objects from two distinct datasets: the upper two objects are from ScanNet200 scenes, while the lower two are from ARKitScenes, demonstrating our framework's adaptability and effectiveness across diverse environments.}
\label{fig:qual}
\vskip -0.15in
\end{figure}

\subsection{Experimental Details}
\label{sec:experimental_details}
In our experiments, we conducted computations using a single RTX 8000 GPU. We utilized posed RGB-D pairs from the ScanNet200 dataset, processing one frame out of every ten frames within the RGB-D sequences. To extract image features from mask crops, we employed the CLIP visual encoder \cite{radford2021learning} from the ViT-L/14 model, known for its feature dimensionality of 768.
For Adaptive Integration, the IoU thresholds were empirically determined to be optimal at $\theta_{2D} = 0.5$ and $\theta_{3D} = 0.9$. 

\begin{table*}[ht]
\centering
\begin{tabular}{@{}cccccccc@{}}
\toprule
3D pathway & 2D pathway & Simple Integration & Dual-path Integration & $\text{AP}$ $\uparrow$ & $\text{AP}_{50}$ $\uparrow$ & $\text{AP}_{25}$ $\uparrow$ \\ 
\midrule
\checkmark  &   &  &  & 16.2 & 20.7 & 22.7 \\
  & \checkmark  &  &  & 10.9 & 15.6 & 20.3 \\
\checkmark  & \checkmark  & \checkmark  &   & 18.2 & 23.8 & 25.8 \\
\checkmark  & \checkmark  &  & \checkmark &  \textbf{19.9} & \textbf{25.0} & \textbf{27.1} \\

\bottomrule
\end{tabular}
\caption{Ablation study on contributions of each component within our Zero-shot Dual-Path Integration Framework in 312 scenes of ScanNet200 \cite{dai2017scannet} validation set.}
\label{table:ab1}
\end{table*}

\begin{table}[ht]
\centering
\begin{tabular}{lccc}
\Xhline{2\arrayrulewidth}
Model & RC $\uparrow$ & $\text{RC}_{50}$ $\uparrow$ & $\text{RC}_{25}$ $\uparrow$ \\ \Xhline{2\arrayrulewidth}
3D pathway   & 14.2 & 19.5   & 22.7    \\ 
2D pathway  & 17.8 & 25.7   & \textbf{32.9}       \\
Dual-Path  & \textbf{21.0} & \textbf{27.9}   & 31.0        \\ \Xhline{2\arrayrulewidth}
\end{tabular}
\caption{Ablation study on the Recall Rate (RC) measured over a range of overlap thresholds, 50\% overlaps, and 25\% overlaps of ``tail'' category in 312 scenes of ScanNet200 \cite{dai2017scannet} validation set.}
\vskip -0.15in
\label{table:ab2}
\end{table}

\subsection{Quantitative Results}
In the comprehensive evaluation presented in Table \ref{table:quan}, performance of our approaches in closed-vocabulary instance segmentation tasks within the ScanNet200 \cite{dai2017scannet} benchmark is provided. 
This distinction in performance is particularly marked within the ``head'' and ``common'' categories, while the disparity narrows in the ``tail'' categories. 

For the previous works on open-vocabulary models, OpenScene \cite{peng2023openscene} is constructed based on 2D model OpenSeg \cite{ghiasi2022scaling} trained on labeled datasets for 2D semantic segmentation. 
OpenMask3D \cite{takmaz2023openmask3d}, a state-of-the-art open-vocabulary model, is built upon the Mask3D for generating class-agnostic 3D mask proposals.
Compared to these previous methods, our Dual-Path Integration Framework has a distinct performance advantage in AP.
This outcome substantiates our initial hypothesis about the efficacy of our method.


\subsection{Qualitative Results and Comparisons}
Fig. \ref{fig:qual} presents the qualitative results that underscore the efficacy of our proposed framework within both seen (ScanNet200) and unseen (ARKitScenes) data, thereby affirming the framework's extensive adaptability and proficiency across diverse environments. 
Being a zero-shot open-vocabulary framework, it enables the segmentation of objects through free-form text queries, even for objects absent in traditional instance segmentation datasets.

\begin{figure*}[h] 
\centering
\includegraphics[width=\textwidth]{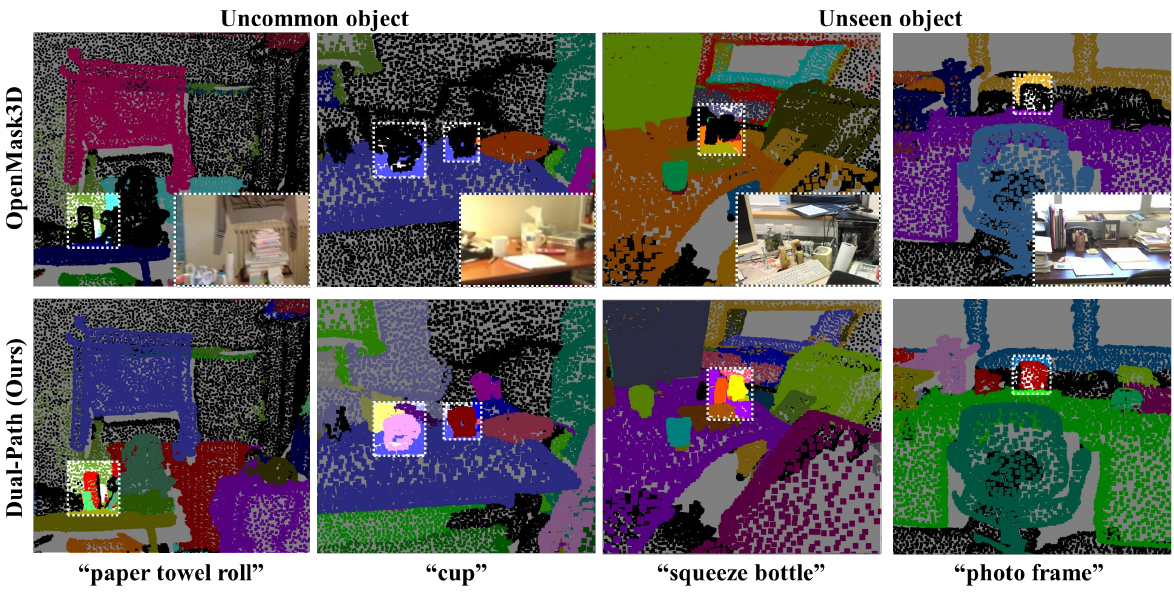} 
\caption{Qualitative comparison between our Dual-Path Integration Framework and OpenMask3D. The black regions indicate no proposals.
Our framework, benefiting from the integration of proposals endowed with high visual understanding capabilities from the 2D pathway, excels in identifying and segmenting uncommon and unseen objects.}
\label{fig:comp}
\end{figure*}

In Figure \ref{fig:comp}, we provide qualitative comparisons that highlight the distinction between our framework and the OpenMask3D in segmenting uncommon objects from the ``tail'' category of ScanNet200, alongside unseen objects not present in the dataset's predefined categories. 
These comparisons underscore our proposed framework's enhanced proficiency in accurately segmenting objects that have posed challenges to previous methods, which predominantly leveraged 3D point cloud data for instance segmentation. 

\begin{table}[ht]
\centering
\begin{tabular}{ccccc}
\Xhline{2\arrayrulewidth}
$\theta_{3D}$ & $\theta_{2D}$ & mAP $\uparrow$ & $\text{AP}_{50}$ $\uparrow$ & $\text{AP}_{25}$ $\uparrow$ \\ \Xhline{2\arrayrulewidth}
0.25            & 0.25            & 18.7          & 24.3             & \textbf{27.2}             \\
0.50            & 0.50            & 19.5          & \textbf{25.0}             & \textbf{27.2}             \\
0.90            & 0.50            & 19.2          & 24.8             & 27.0             \\
0.50            & 0.90            & \textbf{19.9} & \textbf{25.0}    & 27.1    \\
0.90            & 0.90            & 19.8          & 25.0             & 27.0
\\
\Xhline{2\arrayrulewidth}
\end{tabular}
\caption{Ablation study on IoU threshold $\theta_{3D}$ and $\theta_{2D}$ for Adaptive Integration on ScanNet200 \cite{dai2017scannet} validation set.}
\label{table:ab3_threshold}
\vskip -0.15in
\end{table}

\subsection{Ablation Studies}
\label{sec:ablation_studies}
We conducted an ablation study to assess the individual contributions of components within our Dual-Path Integration Framework across 312 scenes from the ScanNet200 validation set, as outlined in Table \ref{table:ab1}.
Our 3D pathway yielded superior performance compared to the reported results in OpenMask3D \cite{takmaz2023openmask3d}, primarily due to our method of directly using the projected 3D bounding box into 2D images to get CLIP-based features for eliminating potential errors in random point selection.
For 2D pathway, relying solely on the 2D data by using 3D projected CLIP \cite{radford2021learning} resulted in suboptimal performance.
While the 'Simple Integration' method of using all proposals from both the 3D and 2D pathways achieves enhanced performance, it falls short of adequately addressing the low quality and redundancy problem of the proposals, failing to leverage the intrinsic strengths unique to each pathway.
In contrast, our Dual-path Integration strategy employs a selective and adaptive approach to integrate the two modalities, yielding improvements over Simple Integration.

In Table \ref{table:ab2}, the recall rate of ``tail'' category of the ScanNet200 validation set is reported.
It highlights the capability of our Dual-Path Integration in identifying uncommon objects. 
A significant factor contributing to the enhancement of the performance is the robust generalization ability offered by the vision-language understanding capabilities inherent in the pre-trained image classifier of the 2D pathway. 
The notable improvement in the recall rate when comparing the Dual-path with the 3D pathway indicates the efficacy of the 2D pathway in our method.

Table \ref{table:ab3_threshold} presents an ablation study on the impact of the IoU thresholds $\theta_{3D}$ and $\theta_{2D}$ within the Adaptive Integration process. The analysis reveals that the Average Precision at a threshold of 25\% ($AP_{25}$) remains relatively unaffected by variations in these thresholds. In contrast, both $AP_{50}$ and the mean Average Precision ($mAP$) exhibit sensitivity to changes in thresholds, suggesting that a higher threshold is crucial for refining segmentation precision. This pattern underscores the importance of carefully calibrating the IoU thresholds to optimize the overall segmentation performance and achieve a balance between recall and precision.


\section{Conclusion}
\label{sec:conclusion}
This paper proposes the Zero-Shot Dual-Path Integration Framework, a model-agnostic strategy that harnesses mask proposals from pretrained models across 3D point cloud and 2D multi-view image modalities for open-vocabulary 3D instance segmentation.
Our novel Conditional Integration process capitalizes on the strengths of instance segmentation within each modality through a two stage methodology: Dual-modality Proposal Matching and Adaptive Integration, aimed at identifying and categorizing significant proposal pairs into distinct categories for effective integration of results from two different modalities. Evaluations conducted on the ScanNet200 benchmark dataset and ARKitScenes dataset illustrate our framework's substantial improvements over prior methodologies, validating the efficacy of integrating 3D and 2D segmentation techniques.

{
    \small
    \bibliographystyle{ieeenat_fullname}
    \bibliography{main}

\begin{thebibliography}{47}
\providecommand{\natexlab}[1]{#1}
\providecommand{\url}[1]{\texttt{#1}}
\expandafter\ifx\csname urlstyle\endcsname\relax
  \providecommand{\doi}[1]{doi: #1}\else
  \providecommand{\doi}{doi: \begingroup \urlstyle{rm}\Url}\fi

\bibitem[Baruch et~al.(2021)Baruch, Chen, Dehghan, Dimry, Feigin, Fu, Gebauer,
  Joffe, Kurz, Schwartz, et~al.]{baruch2021arkitscenes}
Gilad Baruch, Zhuoyuan Chen, Afshin Dehghan, Tal Dimry, Yuri Feigin, Peter Fu,
  Thomas Gebauer, Brandon Joffe, Daniel Kurz, Arik Schwartz, et~al.
\newblock Arkitscenes: A diverse real-world dataset for 3d indoor scene
  understanding using mobile rgb-d data.
\newblock \emph{arXiv preprint arXiv:2111.08897}, 2021.

\bibitem[Chen et~al.(2021)Chen, Fang, Zhang, Liu, and
  Wang]{Chen_HAIS_2021_ICCV}
Shaoyu Chen, Jiemin Fang, Qian Zhang, Wenyu Liu, and Xinggang Wang.
\newblock Hierarchical aggregation for 3d instance segmentation.
\newblock In \emph{ICCV}, 2021.

\bibitem[Cheng et~al.(2022)Cheng, Misra, Schwing, Kirillov, and
  Girdhar]{cheng2021mask2former}
Bowen Cheng, Ishan Misra, Alexander~G. Schwing, Alexander Kirillov, and Rohit
  Girdhar.
\newblock Masked-attention mask transformer for universal image segmentation.
\newblock 2022.

\bibitem[Dai et~al.(2017)Dai, Chang, Savva, Halber, Funkhouser, and
  Nie{\ss}ner]{dai2017scannet}
Angela Dai, Angel~X Chang, Manolis Savva, Maciej Halber, Thomas Funkhouser, and
  Matthias Nie{\ss}ner.
\newblock Scannet: Richly-annotated 3d reconstructions of indoor scenes.
\newblock In \emph{Proceedings of the IEEE conference on computer vision and
  pattern recognition}, pages 5828--5839, 2017.

\bibitem[Ding et~al.(2023)Ding, Yang, Xue, Zhang, Bai, and
  Qi]{ding2022language}
Runyu Ding, Jihan Yang, Chuhui Xue, Wenqing Zhang, Song Bai, and Xiaojuan Qi.
\newblock Pla: Language-driven open-vocabulary 3d scene understanding.
\newblock In \emph{Proceedings of the IEEE/CVF Conference on Computer Vision
  and Pattern Recognition}, 2023.

\bibitem[Engelmann et~al.(2020)Engelmann, Bokeloh, Fathi, Leibe, and
  Nie{\ss}ner]{Engelmann20CVPR}
Francis Engelmann, Martin Bokeloh, Alireza Fathi, Bastian Leibe, and Matthias
  Nie{\ss}ner.
\newblock {3D-MPA: Multi Proposal Aggregation for 3D Semantic Instance
  Segmentation}.
\newblock In \emph{{IEEE Conference on Computer Vision and Pattern Recognition
  (CVPR)}}, 2020.

\bibitem[Engelmann et~al.(2024)Engelmann, Takmaz, Schult, Fedele, Wald, Peng,
  Wang, Litany, Tang, Tombari, et~al.]{engelmann2024opensun3d}
Francis Engelmann, Ayca Takmaz, Jonas Schult, Elisabetta Fedele, Johanna Wald,
  Songyou Peng, Xi Wang, Or Litany, Siyu Tang, Federico Tombari, et~al.
\newblock Opensun3d: 1st workshop challenge on open-vocabulary 3d scene
  understanding.
\newblock \emph{arXiv preprint arXiv:2402.15321}, 2024.

\bibitem[Ghiasi et~al.(2022)Ghiasi, Gu, Cui, and Lin]{ghiasi2022scaling}
Golnaz Ghiasi, Xiuye Gu, Yin Cui, and Tsung-Yi Lin.
\newblock Scaling open-vocabulary image segmentation with image-level labels.
\newblock In \emph{European Conference on Computer Vision}, pages 540--557.
  Springer, 2022.

\bibitem[Graham et~al.(2018)Graham, Engelcke, and Van Der~Maaten]{graham20183d}
Benjamin Graham, Martin Engelcke, and Laurens Van Der~Maaten.
\newblock 3d semantic segmentation with submanifold sparse convolutional
  networks.
\newblock In \emph{Proceedings of the IEEE conference on computer vision and
  pattern recognition}, pages 9224--9232, 2018.

\bibitem[Gu et~al.(2023)Gu, Kuwajerwala, Morin, Jatavallabhula, Sen, Agarwal,
  Rivera, Paul, Ellis, Chellappa, Gan, {de Melo}, Tenenbaum, Torralba, Shkurti,
  and Paull]{conceptgraphs}
Qiao Gu, Alihusein Kuwajerwala, Sacha Morin, {Krishna Murthy} Jatavallabhula,
  Bipasha Sen, Aditya Agarwal, Corban Rivera, William Paul, Kirsty Ellis, Rama
  Chellappa, Chuang Gan, {Celso Miguel} {de Melo}, {Joshua B.} Tenenbaum,
  Antonio Torralba, Florian Shkurti, and Liam Paull.
\newblock Conceptgraphs: Open-vocabulary 3d scene graphs for perception and
  planning.
\newblock \emph{arXiv}, 2023.

\bibitem[Gu et~al.(2021)Gu, Lin, Kuo, and Cui]{gu2021open}
Xiuye Gu, Tsung-Yi Lin, Weicheng Kuo, and Yin Cui.
\newblock Open-vocabulary object detection via vision and language knowledge
  distillation.
\newblock \emph{arXiv preprint arXiv:2104.13921}, 2021.

\bibitem[Ha and Song(2022)]{ha2022semabs}
Huy Ha and Shuran Song.
\newblock Semantic abstraction: Open-world 3{D} scene understanding from 2{D}
  vision-language models.
\newblock In \emph{Proceedings of the 2022 Conference on Robot Learning}, 2022.

\bibitem[He et~al.(2021)He, Shen, and van~den Hengel]{He2021dyco3d}
Tong He, Chunhua Shen, and Anton van~den Hengel.
\newblock {DyCo3d}: Robust instance segmentation of 3d point clouds through
  dynamic convolution.
\newblock In \emph{Proceedings of the IEEE Conference on Computer Vision and
  Pattern Recognition (CVPR)}, 2021.

\bibitem[Hegde et~al.(2023)Hegde, Valanarasu, and Patel]{hegde2023clip}
Deepti Hegde, Jeya Maria~Jose Valanarasu, and Vishal~M Patel.
\newblock Clip goes 3d: Leveraging prompt tuning for language grounded 3d
  recognition.
\newblock \emph{arXiv preprint arXiv:2303.11313}, 2023.

\bibitem[Hou et~al.(2019)Hou, Dai, and Nie{\ss}ner]{hou2019sis}
Ji Hou, Angela Dai, and Matthias Nie{\ss}ner.
\newblock 3d-sis: 3d semantic instance segmentation of rgb-d scans.
\newblock In \emph{Proc. Computer Vision and Pattern Recognition (CVPR), IEEE},
  2019.

\bibitem[Huang et~al.(2023{\natexlab{a}})Huang, Mees, Zeng, and
  Burgard]{huang23vlmaps}
Chenguang Huang, Oier Mees, Andy Zeng, and Wolfram Burgard.
\newblock Visual language maps for robot navigation.
\newblock In \emph{Proceedings of the IEEE International Conference on Robotics
  and Automation (ICRA)}, London, UK, 2023{\natexlab{a}}.

\bibitem[Huang et~al.(2023{\natexlab{b}})Huang, Wu, Chen, Zhao, Zhu, and
  Lasenby]{huang2023openins3d}
Zhening Huang, Xiaoyang Wu, Xi Chen, Hengshuang Zhao, Lei Zhu, and Joan
  Lasenby.
\newblock Openins3d: Snap and lookup for 3d open-vocabulary instance
  segmentation.
\newblock \emph{arXiv preprint arXiv:2309.00616}, 2023{\natexlab{b}}.

\bibitem[Huynh et~al.(2022)Huynh, Kuen, Lin, Gu, and Elhamifar]{huynh2022open}
Dat Huynh, Jason Kuen, Zhe Lin, Jiuxiang Gu, and Ehsan Elhamifar.
\newblock Open-vocabulary instance segmentation via robust cross-modal
  pseudo-labeling.
\newblock In \emph{Proceedings of the IEEE/CVF Conference on Computer Vision
  and Pattern Recognition}, pages 7020--7031, 2022.

\bibitem[Jatavallabhula et~al.(2023)Jatavallabhula, Kuwajerwala, Gu, Omama,
  Chen, Li, Iyer, Saryazdi, Keetha, Tewari, Tenenbaum, {de Melo}, Krishna,
  Paull, Shkurti, and Torralba]{conceptfusion}
{Krishna Murthy} Jatavallabhula, Alihusein Kuwajerwala, Qiao Gu, Mohd Omama,
  Tao Chen, Shuang Li, Ganesh Iyer, Soroush Saryazdi, Nikhil Keetha, Ayush
  Tewari, {Joshua B.} Tenenbaum, {Celso Miguel} {de Melo}, Madhava Krishna,
  Liam Paull, Florian Shkurti, and Antonio Torralba.
\newblock Conceptfusion: Open-set multimodal 3d mapping.
\newblock \emph{arXiv}, 2023.

\bibitem[Jia et~al.(2021)Jia, Yang, Xia, Chen, Parekh, Pham, Le, Sung, Li, and
  Duerig]{jia2021scaling}
Chao Jia, Yinfei Yang, Ye Xia, Yi-Ting Chen, Zarana Parekh, Hieu Pham, Quoc Le,
  Yun-Hsuan Sung, Zhen Li, and Tom Duerig.
\newblock Scaling up visual and vision-language representation learning with
  noisy text supervision.
\newblock In \emph{International conference on machine learning}, pages
  4904--4916. PMLR, 2021.

\bibitem[Jiang et~al.(2020)Jiang, Zhao, Shi, Liu, Fu, and
  Jia]{jiang2020pointgroup}
Li Jiang, Hengshuang Zhao, Shaoshuai Shi, Shu Liu, Chi-Wing Fu, and Jiaya Jia.
\newblock Pointgroup: Dual-set point grouping for 3d instance segmentation.
\newblock \emph{Proceedings of the IEEE Conference on Computer Vision and
  Pattern Recognition (CVPR)}, 2020.

\bibitem[Kerr et~al.(2023)Kerr, Kim, Goldberg, Kanazawa, and Tancik]{lerf2023}
Justin Kerr, Chung~Min Kim, Ken Goldberg, Angjoo Kanazawa, and Matthew Tancik.
\newblock Lerf: Language embedded radiance fields.
\newblock In \emph{International Conference on Computer Vision (ICCV)}, 2023.

\bibitem[Kirillov et~al.(2023)Kirillov, Mintun, Ravi, Mao, Rolland, Gustafson,
  Xiao, Whitehead, Berg, Lo, et~al.]{kirillov2023segment}
Alexander Kirillov, Eric Mintun, Nikhila Ravi, Hanzi Mao, Chloe Rolland, Laura
  Gustafson, Tete Xiao, Spencer Whitehead, Alexander~C Berg, Wan-Yen Lo, et~al.
\newblock Segment anything.
\newblock \emph{arXiv preprint arXiv:2304.02643}, 2023.

\bibitem[Liang et~al.(2023)Liang, Wu, Dai, Li, Zhao, Zhang, Zhang, Vajda, and
  Marculescu]{liang2023open}
Feng Liang, Bichen Wu, Xiaoliang Dai, Kunpeng Li, Yinan Zhao, Hang Zhang,
  Peizhao Zhang, Peter Vajda, and Diana Marculescu.
\newblock Open-vocabulary semantic segmentation with mask-adapted clip.
\newblock In \emph{Proceedings of the IEEE/CVF Conference on Computer Vision
  and Pattern Recognition}, pages 7061--7070, 2023.

\bibitem[Liang et~al.(2021)Liang, Li, Xu, Tan, and Jia]{liang2021instance}
Zhihao Liang, Zhihao Li, Songcen Xu, Mingkui Tan, and Kui Jia.
\newblock Instance segmentation in 3d scenes using semantic superpoint tree
  networks.
\newblock In \emph{Proceedings of the IEEE/CVF International Conference on
  Computer Vision}, pages 2783--2792, 2021.

\bibitem[Liu et~al.(2023)Liu, Zeng, Ren, Li, Zhang, Yang, Li, Yang, Su, Zhu,
  et~al.]{liu2023grounding}
Shilong Liu, Zhaoyang Zeng, Tianhe Ren, Feng Li, Hao Zhang, Jie Yang, Chunyuan
  Li, Jianwei Yang, Hang Su, Jun Zhu, et~al.
\newblock Grounding dino: Marrying dino with grounded pre-training for open-set
  object detection.
\newblock \emph{arXiv preprint arXiv:2303.05499}, 2023.

\bibitem[Lu et~al.(2023)Lu, Chang, Jing, Boularias, and Bekris]{lu2023ovir}
Shiyang Lu, Haonan Chang, Eric~Pu Jing, Abdeslam Boularias, and Kostas Bekris.
\newblock Ovir-3d: Open-vocabulary 3d instance retrieval without training on 3d
  data.
\newblock In \emph{7th Annual Conference on Robot Learning}, 2023.

\bibitem[Mildenhall et~al.(2020)Mildenhall, Srinivasan, Tancik, Barron,
  Ramamoorthi, and Ng]{mildenhall2020nerf}
Ben Mildenhall, Pratul~P. Srinivasan, Matthew Tancik, Jonathan~T. Barron, Ravi
  Ramamoorthi, and Ren Ng.
\newblock Nerf: Representing scenes as neural radiance fields for view
  synthesis.
\newblock In \emph{ECCV}, 2020.

\bibitem[Minderer et~al.()Minderer, Gritsenko, Stone, Neumann, Weissenborn,
  Dosovitskiy, Mahendran, Arnab, Dehghani, Shen, et~al.]{minderer2205simple}
M Minderer, A Gritsenko, A Stone, M Neumann, D Weissenborn, A Dosovitskiy, A
  Mahendran, A Arnab, M Dehghani, Z Shen, et~al.
\newblock Simple open-vocabulary object detection with vision transformers.
  arxiv 2022.
\newblock \emph{arXiv preprint arXiv:2205.06230}.

\bibitem[Ngo et~al.(2023)Ngo, Hua, and Nguyen]{ngo2023isbnet}
Tuan~Duc Ngo, Binh-Son Hua, and Khoi Nguyen.
\newblock Isbnet: a 3d point cloud instance segmentation network with
  instance-aware sampling and box-aware dynamic convolution.
\newblock In \emph{Proceedings of the IEEE/CVF Conference on Computer Vision
  and Pattern Recognition}, pages 13550--13559, 2023.

\bibitem[Peng et~al.(2023)Peng, Genova, Jiang, Tagliasacchi, Pollefeys,
  Funkhouser, et~al.]{peng2023openscene}
Songyou Peng, Kyle Genova, Chiyu Jiang, Andrea Tagliasacchi, Marc Pollefeys,
  Thomas Funkhouser, et~al.
\newblock Openscene: 3d scene understanding with open vocabularies.
\newblock In \emph{Proceedings of the IEEE/CVF Conference on Computer Vision
  and Pattern Recognition}, pages 815--824, 2023.

\bibitem[Radford et~al.(2021)Radford, Kim, Hallacy, Ramesh, Goh, Agarwal,
  Sastry, Askell, Mishkin, Clark, et~al.]{radford2021learning}
Alec Radford, Jong~Wook Kim, Chris Hallacy, Aditya Ramesh, Gabriel Goh,
  Sandhini Agarwal, Girish Sastry, Amanda Askell, Pamela Mishkin, Jack Clark,
  et~al.
\newblock Learning transferable visual models from natural language
  supervision.
\newblock In \emph{International conference on machine learning}, pages
  8748--8763. PMLR, 2021.

\bibitem[Rozenberszki et~al.(2022)Rozenberszki, Litany, and
  Dai]{rozenberszki2022language}
David Rozenberszki, Or Litany, and Angela Dai.
\newblock Language-grounded indoor 3d semantic segmentation in the wild.
\newblock In \emph{European Conference on Computer Vision}, pages 125--141.
  Springer, 2022.

\bibitem[Schult et~al.(2023)Schult, Engelmann, Hermans, Litany, Tang, and
  Leibe]{Schult23ICRA}
Jonas Schult, Francis Engelmann, Alexander Hermans, Or Litany, Siyu Tang, and
  Bastian Leibe.
\newblock {Mask3D: Mask Transformer for 3D Semantic Instance Segmentation}.
\newblock 2023.

\bibitem[Sun et~al.(2022)Sun, Qing, Tan, and Xu]{2211.15766}
Jiahao Sun, Chunmei Qing, Junpeng Tan, and Xiangmin Xu.
\newblock Superpoint transformer for 3d scene instance segmentation, 2022.

\bibitem[Takmaz et~al.(2023)Takmaz, Fedele, Sumner, Pollefeys, Tombari, and
  Engelmann]{takmaz2023openmask3d}
Ay{\c{c}}a Takmaz, Elisabetta Fedele, Robert~W Sumner, Marc Pollefeys, Federico
  Tombari, and Francis Engelmann.
\newblock Openmask3d: Open-vocabulary 3d instance segmentation.
\newblock \emph{arXiv preprint arXiv:2306.13631}, 2023.

\bibitem[VS et~al.(2023)VS, Yu, Xing, Qin, Gao, Niebles, Patel, and
  Xu]{vs2023mask}
Vibashan VS, Ning Yu, Chen Xing, Can Qin, Mingfei Gao, Juan~Carlos Niebles,
  Vishal~M Patel, and Ran Xu.
\newblock Mask-free ovis: Open-vocabulary instance segmentation without manual
  mask annotations.
\newblock In \emph{Proceedings of the IEEE/CVF Conference on Computer Vision
  and Pattern Recognition}, pages 23539--23549, 2023.

\bibitem[Vu et~al.(2022)Vu, Kim, Luu, Nguyen, and Yoo]{vu2022softgroup}
Thang Vu, Kookhoi Kim, Tung~M. Luu, Xuan~Thanh Nguyen, and Chang~D. Yoo.
\newblock Softgroup for 3d instance segmentation on 3d point clouds.
\newblock In \emph{CVPR}, 2022.

\bibitem[Wu et~al.(2023)Wu, Li, Ding, Li, Cheng, Tong, and Loy]{wu2023betrayed}
Jianzong Wu, Xiangtai Li, Henghui Ding, Xia Li, Guangliang Cheng, Yunhai Tong,
  and Chen~Change Loy.
\newblock Betrayed by captions: Joint caption grounding and generation for open
  vocabulary instance segmentation.
\newblock \emph{arXiv preprint arXiv:2301.00805}, 2023.

\bibitem[Wu et~al.(2022)Wu, Shi, Du, Lu, Cao, and Zhong]{wu20223d}
Yizheng Wu, Min Shi, Shuaiyuan Du, Hao Lu, Zhiguo Cao, and Weicai Zhong.
\newblock 3d instances as 1d kernels.
\newblock In \emph{Computer Vision--ECCV 2022: 17th European Conference, Tel
  Aviv, Israel, October 23--27, 2022, Proceedings, Part XXIX}, pages 235--252.
  Springer, 2022.

\bibitem[Xu et~al.(2022)Xu, De~Mello, Liu, Byeon, Breuel, Kautz, and
  Wang]{xu2022groupvit}
Jiarui Xu, Shalini De~Mello, Sifei Liu, Wonmin Byeon, Thomas Breuel, Jan Kautz,
  and Xiaolong Wang.
\newblock Groupvit: Semantic segmentation emerges from text supervision.
\newblock In \emph{Proceedings of the IEEE/CVF Conference on Computer Vision
  and Pattern Recognition}, pages 18134--18144, 2022.

\bibitem[Yang et~al.(2019)Yang, Wang, Clark, Hu, Wang, Markham, and
  Trigoni]{yang2019learning}
Bo Yang, Jianan Wang, Ronald Clark, Qingyong Hu, Sen Wang, Andrew Markham, and
  Niki Trigoni.
\newblock Learning object bounding boxes for 3d instance segmentation on point
  clouds.
\newblock In \emph{Advances in Neural Information Processing Systems}, pages
  6737--6746, 2019.

\bibitem[Yang et~al.(2023)Yang, Wu, He, Zhao, and Liu]{yang2023sam3d}
Yunhan Yang, Xiaoyang Wu, Tong He, Hengshuang Zhao, and Xihui Liu.
\newblock Sam3d: Segment anything in 3d scenes.
\newblock \emph{arXiv preprint arXiv:2306.03908}, 2023.

\bibitem[Yi et~al.(2018)Yi, Zhao, Wang, Sung, and Guibas]{Yi2018GSPNGS}
L. Yi, Wang Zhao, He Wang, Minhyuk Sung, and Leonidas~J. Guibas.
\newblock Gspn: Generative shape proposal network for 3d instance segmentation
  in point cloud.
\newblock \emph{2019 IEEE/CVF Conference on Computer Vision and Pattern
  Recognition (CVPR)}, pages 3942--3951, 2018.

\bibitem[Zhai et~al.(2022)Zhai, Wang, Mustafa, Steiner, Keysers, Kolesnikov,
  and Beyer]{zhai2022lit}
Xiaohua Zhai, Xiao Wang, Basil Mustafa, Andreas Steiner, Daniel Keysers,
  Alexander Kolesnikov, and Lucas Beyer.
\newblock Lit: Zero-shot transfer with locked-image text tuning.
\newblock In \emph{Proceedings of the IEEE/CVF Conference on Computer Vision
  and Pattern Recognition}, pages 18123--18133, 2022.

\bibitem[Zhao et~al.(2023)Zhao, Yan, Yang, Ye, Yang, and Huang]{zhao2023divide}
Weiguang Zhao, Yuyao Yan, Chaolong Yang, Jianan Ye, Xi Yang, and Kaizhu Huang.
\newblock Divide and conquer: 3d point cloud instance segmentation with
  point-wise binarization.
\newblock In \emph{Proceedings of the IEEE/CVF international conference on
  computer vision (ICCV)}, pages 562--571, 2023.

\bibitem[Zhou et~al.(2022)Zhou, Girdhar, Joulin, Kr{\"a}henb{\"u}hl, and
  Misra]{zhou2022detecting}
Xingyi Zhou, Rohit Girdhar, Armand Joulin, Philipp Kr{\"a}henb{\"u}hl, and
  Ishan Misra.
\newblock Detecting twenty-thousand classes using image-level supervision.
\newblock In \emph{European Conference on Computer Vision}, pages 350--368.
  Springer, 2022.

\end{thebibliography}
}


\end{document}